# A Lightweight Attention-based Deep Network via Multi-Scale Feature Fusion for Multi-View Facial Expression Recognition

Ali Ezati, Mohammadreza Dezyani, Rajib Rana, *Member, IEEE*, Roozbeh Rajabi, *Senior Member, IEEE,* and Ahmad Ayatollahi

*Abstract*—Convolutional neural networks (CNNs) and their variations have shown effectiveness in facial expression recognition (FER). However, they face challenges when dealing with high computational complexity and multi-view head poses in real-world scenarios. We introduce a lightweight attentional network incorporating multi-scale feature fusion (LANMSFF) to tackle these issues. For the first challenge, we carefully design a lightweight network. We address the second challenge by presenting two novel components, namely mass attention (MassAtt) and point wise feature selection (PWFS) blocks. The MassAtt block simultaneously generates channel and spatial attention maps to recalibrate feature maps by emphasizing important features while suppressing irrelevant ones. In addition, the PWFS block employs a feature selection mechanism that discards less meaningful features prior to the fusion process. This mechanism distinguishes it from previous methods that directly fuse multi-scale features. Our proposed approach achieved results comparable to state-of-the-art methods in terms of parameter count and robustness to pose variation, with accuracy rates of 90.77% on KDEF, 70.44% on FER-2013, and 86.96% on FERPlus datasets. The code for LANMSFF is available at https://github.com/AE-1129/LANMSFF.

## I. INTRODUCTION

Over the last three decades, facial expression recognition has gained significant attention from researchers due to its powerful and universal nature, particularly in identifying emotions [1]. Ekman and Friesen [2] characterized six basic facial expressions (anger, disgust, fear, happiness, sadness, and surprise) on the basis of cross-cultural research. Their study suggests that humans comprehend particular facial expressions similarly, irrespective of their culture. Each facial expression possesses distinctive features that differentiate it from other expressions, and these features can be effectively acquired and identified by deep learning models. Deep networks have shown more robustness and effectiveness compared to traditional approaches, which often encounter difficulties in generalization to real-world environments [3]. In particular, convolutional neural networks (CNNs) have robust capabilities for extracting, learning, and classifying features. Due to these capabilities, CNNs are highly effective for identifying and discerning subtle variations in facial expressions [4].

However, facial expression recognition (FER) has faced challenges due to the high computational demands of CNNs and the presence of real-world situations. In the literature, few studies have systematically investigated how to reduce the number of trainable parameters in CNNs [5], [6]. A larger number of model parameters requires additional training samples to avoid overfitting [7]. Although early CNN-based methods show promising recognition performance in constrained environments, they often have poorer accuracy in real-world scenarios [8]. Multi-view poses in unconstrained environments may cause alterations in facial areas of images, which complicate accurate classification [9]. Several studies [9], [10], [11], [12], [13], [14], [15] have utilized attention mechanisms and multi-scale features to improve recognition rates in such contexts. For instance, Hu et al. [13] introduced the squeeze-and excitation (SE) module to enhance channel relationships by recalibrating the significance of each channel. Similarly, the convolutional block attention module (CBAM) [12] employs two consecutive stages of channel-wise and spatial-wise attention to refine features. Moreover, Zhao et al. [14] leveraged multi-scale features to extract robust and diverse information.

These challenges necessitate a lightweight model to address the computational demands of CNNs and maintain robustness in multi-view scenarios. In response, we propose a lightweight convolutional attention-based network called LANMSFF, which utilizes two novel modules: mass attention (MassAtt) and point-wise feature selection (PWFS). In MassAtt block, attention maps are generated simultaneously in both spatial and channel dimensions. These attention maps are then multiplied with feature maps to reweight them and highlight critical features while diminishing irrelevant ones. Our approach improves recognition rates by utilizing multi-scale features extracted at different spatial resolutions and levels of abstraction. These features provide robust and varied information in multi-view situations [14]. To this end, we

A. Ezati, MR. Dezyani and A. Ayatollahi are with the School of Electrical Engineering, Iran University of Science and Technology, Iran. E-mail:{aliezati, mohammadreza_dezyani}@elec.iust.ac.ir, ayatollahi@iust.ac.ir
R. Rana is with the School of Mathematics, Physics and Computing, University of Southern Queensland, Australia.
E-mail: Rajib.Rana@unisq.edu.au
R. Rajabi is with the Hyperspectral Imaging Laboratory, University of Alaska Fairbanks, AK, USA.
E-mail: rrajabitoostani@alaska.edu

utilize dilated convolutions and fusion of intermediate features to obtain multi-scale features. Many existing methods simply fuse features from different scales and assign identical weights to these features [14], [15]. However, utilizing all features from diverse perspectives without considering their importance may negatively impact recognition accuracy [16]. Therefore, we introduce the PWFS block to suppress weak components of multi-scale features before fusion. Our major contributions to this work can be summarized as follows:

1) A lightweight model is proposed to strike a balance between accuracy and parameter count. Moreover, this model handles multi-view pose scenarios.
2) MassAtt module is introduced to enable the model to focus on both channel and spatial attention maps simultaneously. By reweighting feature maps, this module highlights critical features while suppressing irrelevant ones.
3) PWFS block is designed to enhance efficiency of the model. This block reduced the number of parameters and selects powerful features before fusing them.
4) Experimental results demonstrate that the proposed method performs comparably to state-of-the-art (SOTA) on various key datasets, including KDEF [17], FER-2013 [18], and FERPlus [19]. Our method shows robustness to multi-view head poses while keeping parameter count optimal.

The remainder of the paper is organized as follows. Section II reviews literature on FER. Next, we describe the proposed models in Section III. Section IV describes the datasets and implementation details, followed by our experimental results. Finally, we conclude the paper in section V by giving insights for future studies.

II. Related Work

*A. Lightweight FER Models*

CNNs have become deeper to achieve accuracy in various applications [4], [7]. However, these deep networks require powerful GPUs due to their high computational complexity. Several recent studies [10], [20] have focused on developing lightweight deep networks to mitigate these challenges.

Shao et al. [21] addressed overfitting and high computational complexity by introducing a CNN architecture. This architecture includes six modules of depth-wise separable residual convolution, followed by a global average pooling (GAP) layer. Each module consists of three depth-wise separable (DWS) [6] convolutional layers, a max pooling layer, and a standard convolution layer within a skip connection path. Gera et al. [20] utilized a lightweight facial recognition model called LightCNN [22] as a pre-trained backbone. They also adopted the efficient channel attention (ECA) mechanism [23] to reduce computational complexity. Jiang et al. [10] introduced the neuron energy-one shot aggregation (NE-OSA) block to reduce parameters. Unlike DenseNet [24], which concatenates all previous features at each succeeding layer, this block aggregates the feature maps of previous convolutional layers only once in the final feature map. In addition, they employed parallel efficient separable convolutions with multi-scale receptive fields on the input image to obtain feature maps at different scales. However, these methods still consist of a large number of parameters, which make them unsuitable for real-world applications. In contrast, we achieve comparable efficacy utilizing a lightweight model with only a small number of parameters.

*B. Visual-attention Based FER*

Attention mechanisms in neural networks concentrate on specific parts of the input while disregarding irrelevant features. Visual attention-based networks have been proposed to localize crucial regions in computer vision tasks, such as FER.

Li et al. [25] proposed a FER system, which utilizes two CNN feature extractors to achieve feature maps from raw and local binary pattern (LBP) source images. Subsequently, these feature maps are utilized by an attention module to construct attention maps. After the attention module refines the feature maps, they are processed by a dense convolutional module comprising four dilated convolutions before a classification module. Wang et al. [9] presented the region attention network (RAN) in which CNN-based backbones capture features from multiple regions in the face. Following this, the framework utilizes two modules, namely self-attention and relation attention, to acquire attention weights for a compact face representation. Huang et al. [11] incorporated grid-wise attention to extract low-level features from distinct facial regions. These extracted features were subsequently fed into a CNN-based network, which collected pyramid feature maps at each convolutional layer and resized them to a fixed-length size. The resized feature maps were then utilized as tokens in a visual transformer, which ultimately extracted global features for FER tasks. Li et al. [26] proposed a strategy that incorporates both local and global features from sliding patches and the entire face, respectively. Subsequently, these features undergo SE attention modules to enhance feature representation. Similarly, Liu et al. [27] proposed a patch attention convolutional vision transformer (PACVT) in order to overcome the challenge of partial occlusion in the FER task. Inspired by CBAM, their backbone incorporates both channel and spatial attention. Gera et al. [20] partitioned the feature maps of a pre-trained CNN-based model into four non-overlapping patches to capture both local and global information. After passing these features through ECA modules to accentuate salient regions, an ensemble learning strategy is employed to enhance overall performance. Jiang et al. [10] introduced multi-path interactively squeeze and extraction attention (MPISEA) module to divide channel-wise feature maps from the stem module into two patches. These patches undergo linear transformations to generate subspaces. Following the application of dot products to these subspaces, a sigmoid function generates an attention mask, which is finally multiplied with the input image. Due to the effectiveness of visual attention in these methods, we use attention mechanisms in both early and later blocks to incorporate low and high-level features into attention-based representations.

TABLE I
COMPARATIVE SUMMARY OF OUR STUDY AND PREVIOUS LITERATURE

| Paper/Author (Year) | Method | Lightweight | Attention | | Multi-scale fusion | | Multiview Scenario |
|---|---|---|---|---|---|---|---|
| | | | Channel | Spatial | Direct | Constraint | |
| Shao et al.[21] (2019) | CNN | ✓ | ✗ | ✗ | ✗ | ✗ | ✗ |
| Wang et al. [9] (2020) | RAN | ✗ | ✗ | ✓ | ✗ | ✗ | ✓ |
| Li et al. [26] (2020) | SPWFA-SE | ✗ | ✓ | ✗ | ✓ | ✗ | ✗ |
| Liu et al. [28] (2021) | DML-Net | ✗ | ✗ | ✗ | ✓ | ✗ | ✓ |
| Huang et al. [11] (2021) | FER-VT | ✗ | ✗ | ✓ | ✗ | ✗ | ✗ |
| Gera et al.[20] (2022) | CERN | ✓ | ✓ | ✗ | ✗ | ✗ | ✓ |
| Liu et al.[27] (2023) | PACVT | ✗ | ✓ | ✓ | ✓ | ✗ | ✗ |
| Jiang et al. [10] (2023) | TST-RRN | ✓ | ✓ | ✓ | ✓ | ✗ | ✗ |
| Xiao et al.[16] (2023) | CFNet | ✗ | ✗ | ✗ | ✗ | ✓ | ✗ |
| Ma et al. [29] (2023) | VTFF | ✗ | ✓ | ✓ | ✗ | ✓ | ✓ |
| This paper (2024) | LANMSFF | ✓ | ✓ | ✓ | ✓ | ✓ | ✓ |

*C. Multi-scale Fusion*

The complementary attributes of multi-perspective features have motivated several studies to integrate these features.

Li et al. [26] extracted local and global features from sliding patches and entire facial images, respectively. They fused these two types of features in a direct manner to obtain a synergistic outcome. Liu et al. [27] leveraged both global and local features by directly merging them via concatenation. After extracting feature maps using a CNN backbone, local features were obtained via a vision transformer to capture long-range dependencies between patches. Jiang et al. [10] utilized several parallel separable convolutions with multi-scale receptive fields to obtain multi-scale feature maps, which were fused through element-wise addition. In contrast to the above-mentioned methods that adopted direct feature fusion, only a few approaches consider the importance of individual features before fusion. Xiao et al. [16] assigned adaptive weights to features based on their significance before fusing to enhance recognition rates in unconstrained conditions. Ma et al. [29] introduced attentional selective fusion (ASF) module to adaptively fuse LBP and CNN feature. ASF prioritizes discriminative features by assigning them greater weight, with fusion weights generated through a global-local attention mechanism. However, our work employs both direct fusion and constrained fusion. The latter is achieved by eliminating redundant features prior to the fusion process.

*D. Multi-view Scenarios*

Real-world images of human faces are not always captured in frontal positions, which poses challenges for FER tasks since important facial regions may disappear under such conditions [30]. Liu et al. [28] employed three ResNet50 [31] models to extract features from distinct facial areas (eyes, mouth) as well as the entire face. These features were subsequently fused and mapped into three embedding features for a multi-task learning process. This process addressed the tasks of FER, pose estimation, and embedding distances. RAN [9] and CERN [20] proposed end-to-end learning models based on attention mechanism, wherein salient regions of facial images were emphasized, resulting in pose-robust models. Ma et al. [29] proposed visual transformers with feature fusion (VTFF) model that exhibits promising generalization under pose-variant conditions. Following extraction and fusion of LBP and RGB features, VTFF transformed the fused features into sequences of visual words. These sequences were then utilized in a transformer architecture to learn the relationship between them. All of these approaches exhibit a suitable degree of robustness to pose variation. Accordingly, we utilize an attention mechanism and feature selection to enhance the robustness against pose variation. Table I presents a brief comparison between existing literature and our work, categorizing them into four broad categories: lightweight FER models, attention mechanisms, multi-scale fusion, and multi-view scenarios.

III. PROPOSED METHOD

*A. Architecture Overview*

We design LANMSFF incorporating two novel blocks, MassAtt and PWFS, to achieve high accuracy while minimizing the number of parameters. MassAtt enables the model to focus on the most important regions of images, while PWFS removes less significant features. The model comprises four sequential blocks that share similarities between the first and third blocks, as well as between the second and fourth blocks (see Fig. 1).

The first and third blocks each initiate with three convolutional layers, with 66 channels in the first block and 78 channels in the third block. The first and third convolutional layers in these blocks are standard convolutions, whereas the second layer is a DWS convolution to reduce computational complexity. After these convolutions, batch normalization (BN) is utilized to normalize activations and reduce overfitting. Subsequently, a max-pooling layer with a 2×2 kernel size and a stride of 2 is employed to decrease spatial dimensions. Finally, a dropout layer is introduced to enhance the learning of robust features by randomly removing some neurons during the training process.

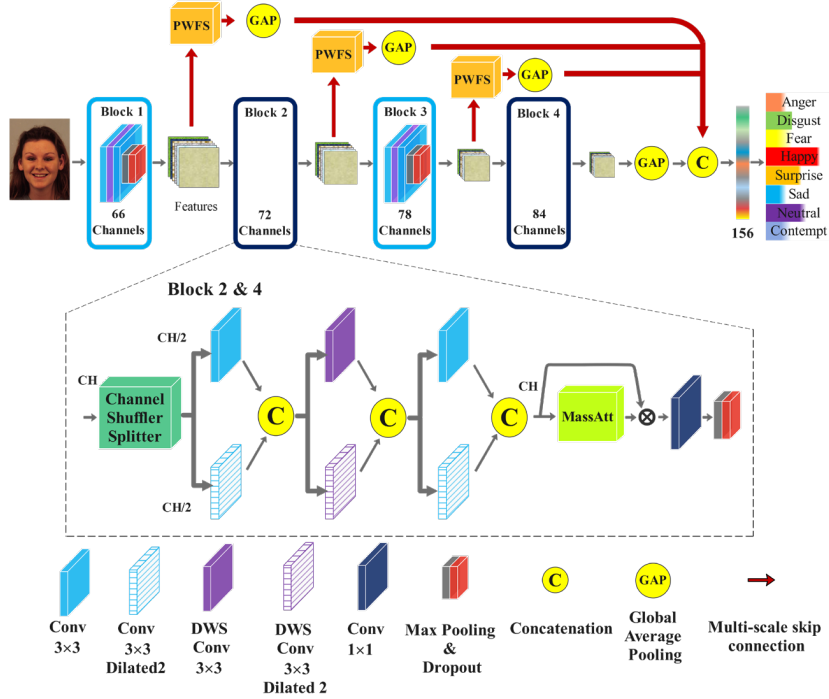

**Fig. 1.** The overview of our proposed model

The second block (with 72 channels) and the fourth block (with 84 channels) utilize a group convolution structure to reduce the number of parameters by performing separate convolutions on divided input channels. These blocks begin with the shuffling and splitting module. This module adds randomness and divides the input feature maps into two groups based on channels. The Shuffling operation rearranges the order of channels in feature maps before separating them into groups. This block facilitates information flow across channels within both groups and mitigates the isolation of information within a single group caused by group convolutions [32]. Then, the divided feature maps are directed into two parallel convolutional paths. One path employs three 3×3 convolutions without dilation, whereas the second one utilizes three 3×3 dilated convolutions with a dilation rate of two. Dilated convolutions are effective for capturing multi-scale information by introducing gaps between kernel elements. This approach expands the receptive field without reducing resolution or introducing additional parameters [33]. Capturing information at different scales by combining standard and dilated convolutions facilitates the extraction of multi-scale features, which provide a comprehensive understanding of input data. Additionally, incorporating DWS convolutions in each path reduces the number of parameters. Features from both paths are concatenated after each convolution to integrate information across the two parallel groups, resulting in more detailed and diverse feature representations. Following this, a series of operations is applied on the outputs of the two branches: concatenation, a MassAtt block, multiplication, a 1×1 convolutional layer, batch normalization (BN), max pooling, and a dropout layer.

The multi-scale features from the outputs of the first three blocks pass through PWFS blocks, which remove the weakest one-third of features at each scale. Subsequently, GAP operations reduce the dimension of feature maps, preparing them for concatenation. The results of the fourth block and the GAP operations are then concatenated to generate a feature vector with 156 nodes. Finally, this vector is fed into a classification layer with neurons equal to the number of classes, which maps the extracted features to the corresponding expression label. Notably, the convolutional layers utilize rectified linear unit (ReLU) activation function, defined as $f(z) = \max(z, 0)$. Additionally, the classification layer computes the probability of each class using softmax activation function, as follows:

$$\text{softmax}(z_i) = \frac{\exp(z_i)}{\sum_0^{n-1} \exp(z_i)} \quad (1)$$

where $z_i$ is *i*-th class score, and *n* is the number of classes.

*B. PWFS Block*

Numerous existing methods have employed fusion to leverage multi-scale features by assigning these features uniform weights [14], [15]. Fundamentally, the features extracted for direct fusion may contain redundant information that obscures crucial facial details. This issue becomes particularly evident in multi-view FER and may deteriorate the performance of FER systems due to the inclusion of background regions and unimportant features, which contain redundant details [34].

Inspired by the max feature map (MFM) approach [22], which suppresses low-activation neurons by selecting the neurons with maximum and median values from feature maps. We design a refined method, called the PWFS block, by eliminating less informative features from each scale before fusion. This approach minimizes the parameter count and computations, resulting in a more efficient and compact architecture. Specifically, our PWFS block enhances the MFM process by averaging the maximum and median values. The selection of maximum values highlights the strongest signals but may emphasize noise or outliers. By incorporating an averaging operation, the PWFS block suppresses noise, thereby improving noise resistance.

As depicted in Fig. 2, the process of the PWFS block is initiated by examining the input feature maps denoted as $f = \{1, ..., C\} \in R^{(H \times W)}$, where $C$, $H$, and $W$ represent the number of channels, height, and width, respectively. Subsequently, the feature map is divided into three distinct sub-groups based on channels: $S0, S1, and\ S2 \in R^{(\frac{C}{3} \times H \times W)}$. This division into three sub-groups, inspired by MFM, balances between computational efficiency and feature discrimination. Following this, the block preserves the highest and middle features as well as discarding the lowest ones across identical positions in the three sub-group feature maps. The two remaining sub-groups are then squeezed into a single sub-group through a point-wise average process. This process involves averaging the values of corresponding elements at the same spatial position across the two sub-groups. Notably, this process preserves the spatial dimensions (i.e., height $H$ and width $W$) of the feature maps. The mathematical expression of this block is as follows:

$$v_{chw} = \frac{1}{2}\left(\max\left(x_{chw}^{s0}, x_{chw}^{s1}, x_{chw}^{s2}\right) + mdn\left(x_{chw}^{s0}, x_{chw}^{s1}, x_{chw}^{s2}\right)\right) \quad (2)$$

where $X_{chw}$ represents an element located at position $(h, w)$ within the c-th feature map of each respective sub-group, while $v_{chw}$ corresponds to the resulting output of PWFS. The operation $mdn$ denotes the median process.

*C. MassAtt Block*

The MassAtt block applies dual-path attention mechanisms simultaneously to focus on the most relevant channels and spatial regions, which is crucial for pose-variant FER. In the proposed MassAtt block, we develop a channel attention mechanism to assign weights to each channel based on its significance for the FER task, while a parallel spatial attention mechanism is integrated into the attention block to identify the most important regions of the facial feature map. This hybrid attention approach reduces the number of parameters compared to computing a full attention map directly. In addition, this structure ensures that each path is optimally configured for its specific role. The combination of simultaneous channel and spatial attention through multiplication enables a synergistic interaction.

As illustrated in Fig. 3, the schematic input-output sequence of MassAtt consists of two parallel paths. The upper attention path comprises a GAP layer and a bottleneck structure that generates a channel attention map, akin to SE attention mechanism [13]. The GAP operation generates channel descriptors by condensing each channel's global spatial context into a single value, which is effective for pose-variant FER where changes in head poses might shift facial regions. The descriptors are then applied to a two-layer bottleneck MLP structure to enhance computational efficiency while retaining crucial features. The first layer reduces dimensionality by a factor of 4, which is chosen based on empirical observations. This reduction enables the model to prioritize the most relevant channel features and discards less critical ones. The second layer restores dimensionality to the original input channel. These layers allow the model to learn which channels are more significant based on inter-channel dependencies.

In the same way, the lower attention path generates the spatial attention map, which allows the model to focus on important regions rather than all regions. This path initiates with the calculation of a 2D spatial descriptor along the channel axis by averaging the input feature maps in a point-wise GAP operation. Subsequently, two convolutional layers with a stride of two are used to reduce the spatial dimensions and achieve a compressed understanding of the spatial layout. Convolutional layers are robust to spatial variations, which is critical in pose-variant FER. Following this, two transposed convolutional layers are used to re-expand the spatial dimensions to the original input size, while focusing on important spatial regions. In the final stage, the channel and spatial attention maps are multiplied to form the mass attention map, which preserves the original input dimensions. This attention fusion enables the model to focus on critical regions and channels under different head poses. Finally, a sigmoid activation function is applied to rescale the values between 0 and 1. This normalization of weights allows the model to suppress irrelevant features while amplifying significant ones. Mathematically, MassAtt block can be defined as follows:

$$\begin{cases} A_m = \sigma(A_c \otimes A_s) \\ A_c = (Z_1 \delta(Z_0 D_c)) \\ A_s = (Z_5 \delta(Z_4 \delta(Z_3 \delta(Z_2 D_s)))) \end{cases} \quad (3)$$

where $A_m$ represents the mass attention and $A_c \in \mathbb{R}^{C \times 1 \times 1}$ and $A_s \in \mathbb{R}^{1 \times H \times W}$ denote the channel and spatial attention maps. Here $\otimes$, $\sigma$, and $\delta$ represent element-wise multiplication, sigmoid, and ReLU functions, respectively. The weights of dense layers are denoted as $Z_0 \in \mathbb{R}^{(C/r) \times C}$, $Z_1 \in \mathbb{R}^{C \times (C/r)}$. The weights of convolutional layers are represented by $Z_2 \in \mathbb{R}^{2 \times 1 \times 3 \times 3}$ and $Z_3 \in \mathbb{R}^{4 \times 2 \times 3 \times 3}$, while the weights of transposed convolutional layers are denoted as $Z_4 \in \mathbb{R}^{4 \times 4 \times 3 \times 3}$ and $Z_5 \in \mathbb{R}^{1 \times 4 \times 3 \times 3}$. Moreover, $D_c$ and $D_s$ are channel and spatial descriptors are defined as follows:

$$\begin{cases} D_c = \frac{1}{H \times W} \sum_{i=1}^{H} \sum_{j=1}^{w} X(i,j)_k & k=1,2,...,C \\ D_s = \frac{1}{C} \sum_{k=1}^{C} X(k)_{i,j} & i=1,2,...,H, \ j=1,2,...,W \end{cases} \quad (4)$$

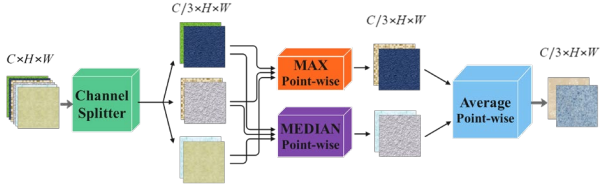

**Fig. 2.** The illustration of PWFS block.

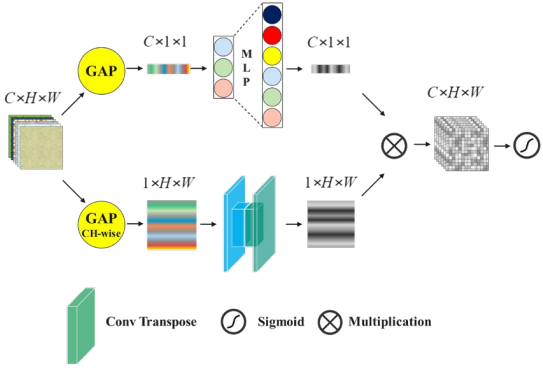

**Fig. 3.** The structure of our MassAtt module.

IV. EXPERIMENTS

We conduct a series of experiments to evaluate and compare the performance of the proposed model. For this purpose, we employ three well-known FER datasets, namely KDEF, FER-2013 and FERPlus.

*A. Datasets*

*KDEF* [17] contains 4900 RGB images with corresponding labels, which have balanced distribution among each expression of disgust, anger, fear, happy, surprise, sad, and neutral. Each expression was performed by 70 actors at five different multi-view angles (-90°, -45°, 0°, 45° and 90°). Some examples are illustrated in Fig. 4.

*FER-2013* [18] was introduced in ICML-2013 challenge. The Google image search API was employed to find facial images using a set of 184 keywords that are related to emotions. After rejecting the incorrectly labeled images, each image was cropped and resized to 48×48 pixels. FER-2013 contains 35887 grayscale images with corresponding labels, distributed as 4953 angry, 547 disgust, 5121 fear, 8989 happy, 4002 surprise, 6077 sad, and 6198 neutral. The dataset consists of 28,709 training samples and 7,178 test samples, equally divided between validation and test sets.

*FERPlus* [19] is a modified version of FER-2013, where the labels were improved using crowdsourcing. Therefore, 10 taggers were asked to label the images into eight classes: neutral (Ne), happy (Ha), surprise (Su), sad (Sa), angry (An), disgust (Di), fear (Fe), and contempt (Co). It is notable that we consider the majority label for each sample as a final label. The dataset consists of 25,371 training samples, 3,225 validation samples, and 3,160 private test images. Samples of FER-2013 and FERPlus are shown in Fig. 5.

We evaluate our model on *Pose-FERPlus* dataset to demonstrate its robustness under varying poses. This dataset is a subset of FERPlus collected by [9] including samples with poses greater than 30° and 45°. In addition, we utilize the indices provided by [9] for samples with poses larger than 30° and 45° to create the *Pose-FER-2013* dataset.

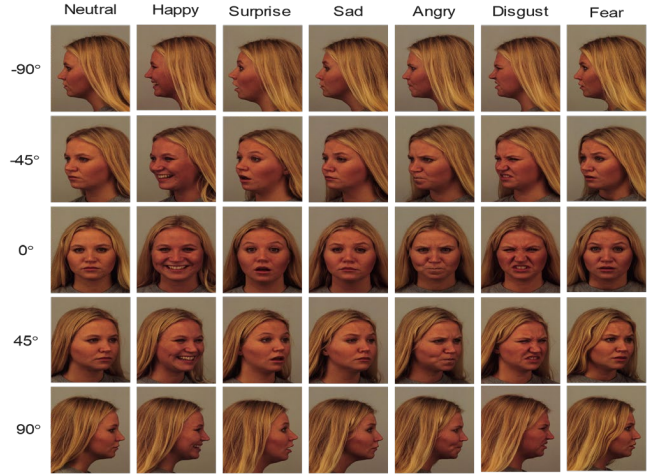

**Fig. 4.** Samples of facial expression images from KDEF dataset.

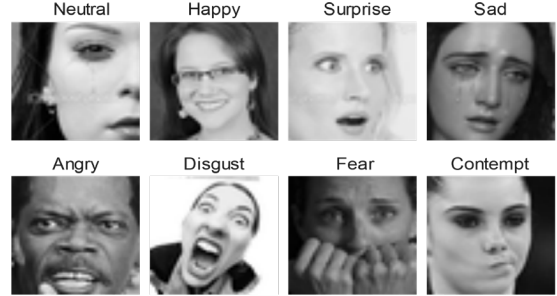

**Fig. 5.** Samples of facial expression images from FER-2013 [18] and FERPlus [19].

*B. Experimental Setup*

The input images were first resized to 64×64 pixels and normalized using min-max normalization. Subsequently, three data augmentation techniques (image cropping, rotation, and flipping) were employed to expand the training set, resulting in three supplementary synthetic 64×64 images for each original image.

We perform 5-fold cross-validation on the KDEF dataset since the publishers do not provide the test set. The proposed model was implemented using the TensorFlow framework, and all experiments were conducted in Google Colab environment. The model was compiled with categorical cross-entropy loss and a batch size of 32. The weights between neurons were updated using the Adam optimizer. The initial learning rate was set to 0.001 and decayed after every eight epochs if the validation loss failed to improve.

Multiple evaluation metrics were employed to assess the performance of the model, including parameter count, accuracy (Acc), information density (ID) [35], and accuracy variance across different poses (Var). The metric ID is defined as the accuracy-to-parameter ratio (in millions). This metric indicates the degree to which a model employs its parameters efficiently to achieve a specific level of accuracy. It is useful to compare models with different sizes. Therefore, an architecture with high ID maintains high accuracy while having fewer parameters. We utilize Var to evaluate the robustness of the model to pose variations, unlike previous literature that considers only overall accuracy. This metric represents the variance in accuracy across different poses, where a lower Var value indicates consistent performance regardless of pose variations. A robust method for pose variation shows a slight degradation in accuracy for various poses, resulting in a low Var value.

*C. Results and Discussion*

1) *Experiments on KDEF*:

We conduct an experiment on the KDEF dataset to assess the performance of LANMSFF in multi-view situations. Table II presents LANMSFF's recognition rate for each expression under multi-view scenarios across five poses. As shown in this Table, our proposed model achieves an accuracy of 90.77% on the entire dataset. The robustness of the model against pose variation is clearly demonstrated by the negligible difference between the overall accuracy and the accuracy of each pose. These findings are in line with other studies [28], [36], which found that the 0° pose achieves the highest accuracy and the -90° pose has the lowest accuracy. A notable increase in accuracy is observed for "Neutral" and a significant decrease in accuracy is noted for "Disgust" when the viewpoint shifts from frontal to non-frontal. This observation implies that the model might struggle to capture critical facial features essential for accurate expression recognition in non-frontal views. Specifically, as the variation in angles increases, the model's performance decreases. Table III displays the confusion matrix for the KDEF dataset, showing that "Happy" and "Neutral" achieve high accuracy, whereas "Fear" exhibits lower accuracy.

TABLE II
RECOGNITION ACCURACIES FOR MULTI-VIEW IMAGES ON KDEF DATASET

| Pose | ACC (%) | | | | | | | |
|---|---|---|---|---|---|---|---|---|
| | Whole | An | Di | Fe | Ha | Sa | Su | Ne |
| -90° | 89.44 | 87.02 | 82.52 | 82.92 | 99.26 | 85.03 | 90.78 | 98.57 |
| -45° | 91.18 | 89.97 | 88.33 | 81.46 | 97.83 | 89.40 | 92.87 | 98.59 |
| 0° | 92.04 | 89.84 | 97.06 | 80.62 | 99.31 | 88.36 | 92.79 | 95.71 |
| 45° | 91.00 | 88.53 | 90.55 | 79.91 | 99.26 | 87.87 | 92.12 | 98.54 |
| 90° | 90.17 | 89.28 | 88.50 | 77.92 | 97.88 | 85.74 | 95.71 | 96.47 |
| Overall | 90.77 | 88.93 | 89.39 | 80.57 | 98.70 | 87.28 | 92.85 | 97.58 |

TABLE III
THE NORMALIZED CONFUSION MATRIX ON KDEF DATASET

| | | Predict | | | | | | |
|---|---|---|---|---|---|---|---|---|
| | | Fe | An | Di | Ha | Ne | Sa | Su |
| Actual | Fe | **80.60** | 1.69 | 1.46 | 1.29 | 1.15 | 6.87 | 6.94 |
| | An | 1.99 | **88.95** | 4.34 | 1.57 | 1.28 | 1.87 | 0 |
| | Di | 2.46 | 3.18 | **89.44** | 1.59 | 0.15 | 3.18 | 0 |
| | Ha | 0.43 | 0.14 | 0.57 | **98.72** | 0.14 | 0 | 0 |
| | Ne | 0.15 | 0.71 | 0 | 0 | **97.57** | 1.57 | 0 |
| | Sa | 4.13 | 1.44 | 2.57 | 0.57 | 3.82 | **87.33** | 0.14 |
| | Su | 6.14 | 0 | 0 | 0.15 | 0.72 | 0.29 | **92.70** |

2) *Experiments on FER-2013*:

As shown in Table IV, LANMSFF achieves an average accuracy of 70.44% across the entire FER-2013 test set. However, the accuracy decreases for test set images with non-frontal views. These observed difficulties in recognition rate can be attributed to the challenging nature of the in-the-wild images in the FER-2013 dataset. Table V displays the confusion matrix for the FER-2013, indicating a low recognition rate for the "Disgust" and "Neutral" expressions. The highest levels of misclassification occur between "Disgust" and "Neutral", as well as between "Neutral" and "Surprise". This issue can be justified by false tags, non-facial samples, and imbalanced classes in this dataset. Table VI demonstrates that our model outperforms other methods (e.g., [21], [37], [38], [39], [40], [41], [42], [43]) on the FER-2013 dataset, particularly when considering the ID metric. Our LANMSFF achieves a notable ID by significantly reducing the number of parameters, albeit with a slight trade-off in accuracy compared to existing approaches that report higher accuracy. For instance, although AMP_NET [43] achieves 4% higher accuracy than our model, it uses 276 times more parameters, resulting in a lower ID.

TABLE IV
RECOGNITION RATE UNDER MULTI-VIEW POSE ON FER-2013 AND POSE-FER-2013 DATASETS

| Pose | ACC (%) | | | | | | | |
|---|---|---|---|---|---|---|---|---|
| | Whole | An | Di | Fe | Ha | Sa | Su | Ne |
| >30° | 69.15 | 63.74 | 75.00 | 37.33 | 88.89 | 55.71 | 82.20 | 76.00 |
| >45° | 66.82 | 59.34 | 75.00 | 37.97 | 85.21 | 59.31 | 82.14 | 71.55 |
| Overall | 70.44 | 62.12 | 74.55 | 45.08 | 89.42 | 54.71 | 80.53 | 79.55 |

TABLE V
THE NORMALIZED CONFUSION MATRIX ON FER-2013

| | | Predict | | | | | | |
|---|---|---|---|---|---|---|---|---|
| | | Fe | An | Di | Ha | Ne | Sa | Su |
| Actual | Fe | **62.12** | 1.43 | 8.35 | 3.25 | 12.42 | 1.43 | 11.00 |
| | An | 18.17 | **74.55** | 0 | 0 | 1.82 | 3.64 | 1.82 |
| | Di | 13.06 | 1.14 | **45.08** | 2.65 | 17.05 | 8.71 | 12.31 |
| | Ha | 1.02 | 0 | 0.35 | **89.42** | 2.39 | 2.50 | 4.32 |
| | Ne | 9.09 | 0.51 | 8.75 | 4.21 | **54.71** | 1.18 | 21.55 |
| | Sa | 2.64 | 0.48 | 7.93 | 4.58 | 1.44 | **80.53** | 2.40 |
| | Su | 3.83 | 0.16 | 2.40 | 3.51 | 9.11 | 1.44 | **79.55** |

TABLE VI
COMPARISON OF ACCURACY (%) AND PARAMETERS COUNTS BETWEEN PROPOSED METHOD AND PREVIOUS WORKS ON FER-2013

| Methods | Acc (%) | Params | ID |
|---|---|---|---|
| SHCNN [37] | 69.10 | 8700K | 7.9 |
| Light CNN [21] | 68.00 | 1108K | 61.3 |
| BReG-NeXt-32 [38] | 69.11 | 1900K | 36.37 |
| BReG-NeXt-50 [38] | 71.53 | 3100K | 23.07 |
| Dense_FaceLiveNet [39] | 69.99 | 15300K | 4.57 |
| Block-FreNet [40] | 64.41 | 9000K | 7.16 |
| SAN-CNN [41] | 74.17 | 6580K | 11.2 |
| AR-TE-CATFFNet [42] | 74.84 | 32117K | 2.3 |
| AMP_NET [43] | 74.48 | 105670K | 0.7 |
| LANMSFF | 70.44 | 358K | 196 |

3) *Experiments on FERPlus*:

The performance of LANMSFF on FERPlus is presented in Table VII, demonstrating an average accuracy of 86.96% across the entire test dataset. However, the accuracy of the model decreases when confronted with images having >30° and >45° poses, consistent with the results seen in existing approaches [9], [20], [29], [43], [46]. Table VIII depicts the confusion matrix for FERPlus. The results indicate that "Neutral" and "Happiness" demonstrate the most accurate prediction, whereas "Contempt", "Disgust", and "Fear" exhibit inaccuracies. Insufficient samples for "Contempt", "Disgust", and "Fear" make these three classes more difficult to distinguish and prone to misclassification as "Neutral", "Angry", and "Surprise", respectively. Considering the discrepancies in prediction accuracy across these classes, the class imbalance distribution cannot be overlooked. The number of samples directly affects the model's ability to learn features, since there is a positive correlation between the large quantity of "Neutral" and "Happiness" samples and their high accuracy. Table IX depicts the comparison of our model with SOTA methods [44], [10]. Our results indicate that a substantial reduction in model parameters results in a significant ID, albeit with a slight decrease in accuracy, compared to certain methods. As an illustration, our accuracy is marginally 3% lower compared to FENN [44] and PACVT [27], which use 30 and 74 times more parameters than our model, respectively.

which utilizes both an attention mechanism and dynamic constraint multi-task learning. OCA-MTL [36] achieves an accuracy of 89.04% using both frontal and non-frontal images as inputs for a Siamese network, which is 1.73% lower than the accuracy of our method. In contrast to our approach, OCA-MTL relies only on channel attention to focus on salient facial regions. Our Var is 0.66, compared to 3.69 for DML-NET [28], 5.42 for SSA-Net [45], and 2.74 for OCA-MTL [36]. The lower Var value confirms the robustness of our approach when applied to non-frontal images.

Our work is compared with SOTA methods in Table XI, which includes CERN [20], RAN [9], AMP_NET [43], VTFF [29], and FG-AGR [46] on Pose-FERPlus. This table computes Var for Pose $\geq 30°$, Pose $\geq 45°$, and the entire dataset to evaluate the robustness of each method to head-pose variations. Our model achieves a Var value of 1.13, compared to 1.88 for CERN, 2.31 for FG-AGR, and 14.2 for RAN methods. The lower Var value of our model demonstrates greater robustness to head-pose variations compared to prior studies.

TABLE VII
RECOGNITION RATE UNDER MULTI-VIEW POSE ON FERPLUS AND POSE-FERPLUS DATASETS

| Pose | ACC (%) | | | | | | | | |
|---|---|---|---|---|---|---|---|---|---|
| | Whole | Ne | Ha | Su | Sa | An | Di | Fe | Co |
| >30° | 86.92 | 90.34 | 94.41 | 88.98 | 72.85 | 86.27 | 20.00 | 50.00 | 0 |
| >45° | 84.68 | 88.19 | 93.62 | 88.14 | 69.66 | 80.77 | 100 | 50.00 | 0 |
| Overall | 86.96 | 89.38 | 94.96 | 89.09 | 71.99 | 85.50 | 43.75 | 55.81 | 13.33 |

TABLE VIII
THE NORMALIZED CONFUSION MATRIX ON FERPLUS

| | | Predict | | | | | | | |
|---|---|---|---|---|---|---|---|---|---|
| | | Ne | Ha | Su | Sa | An | Di | Fe | Co |
| Actual | Ne | 89.38 | 2.12 | 1.11 | 6.28 | 0.74 | 0.09 | 0.28 | 0 |
| | Ha | 2.46 | 94.96 | 1.35 | 1.01 | 0.22 | 0 | 0 | 0 |
| | Su | 4.31 | 1.78 | 89.09 | 1.02 | 1.52 | 0 | 2.28 | 0 |
| | Sa | 20.68 | 2.36 | 0.78 | 71.99 | 3.14 | 0 | 1.05 | 0 |
| | An | 5.95 | 4.46 | 1.49 | 1.86 | 85.50 | 0.37 | 0.37 | 0 |
| | Di | 6.25 | 6.25 | 12.5 | 0 | 31.25 | 43.75 | 0 | 0 |
| | Fe | 3.48 | 0 | 27.91 | 10.47 | 2.33 | 0 | 55.81 | 0 |
| | Co | 46.67 | 6.67 | 0 | 20 | 13.33 | 0 | 0 | 13.33 |

TABLE IX
COMPARISON OF ACCURACY (%) AND PARAMETERS COUNTS BETWEEN PROPOSED METHOD AND PREVIOUS WORKS ON FERPLUS

| Methods | Acc (%) | Params | ID |
|---|---|---|---|
| SHCNN [37] | 86.50 | 8700K | 9.94 |
| FENN [44] | 89.53 | 11470K | 7.80 |
| PACVT [27] | 88.72 | 28400K | 3.1 |
| TST-RRN [10] | 89.64 | 1820K | 49.25 |
| RAN [9] | 89.16 | 11180K | 7.94 |
| CERN [20] | 88.17 | 1450K | 60.8 |
| VTFF [29] | 88.81 | 80000K | 1.1 |
| LANMSFF | 86.96 | 358K | 242 |

4) *Robustness Analysis to multi-view poses*:
Table X demonstrates that our proposed model achieves higher accuracy than SOTA methods, including DML-NET [28], which employs dynamic constraint multi-task learning to estimate head poses as an additional task, and SSA-Net [45],

TABLE X
COMPARISON OF ACCURACIES BETWEEN THE PROPOSED METHOD AND PREVIOUS WORKS ON KDEF

| Methods | Acc (%) | | | | | | Var |
|---|---|---|---|---|---|---|---|
| | -90° | -45° | 0° | +45° | +90° | Overall | |
| DML-NET [28] | 89.20 | 88.80 | 91.30 | 85.60 | 86.10 | 88.20 | 3.69 |
| SSA-Net [45] | 90.20 | 90.20 | 89.70 | 83.50 | 89.10 | 88.50 | 5.42 |
| OCA-MTL [36] | 87.04 | 89.18 | 92.24 | 89.18 | 87.55 | 89.04 | 2.74 |
| LANMSFF | 89.44 | 91.18 | 92.04 | 91.00 | 90.17 | 90.77 | 0.66 |

TABLE XI
COMPARISON OF ACCURACIES BETWEEN THE PROPOSED METHOD AND PREVIOUS WORKS ON POSE-FERPLUS

| Methods | Acc (%) | | | Var |
|---|---|---|---|---|
| | Pose $\geq 30°$ | Pose $\geq 45°$ | Overall | |
| RAN [9] | 82.23 | 80.40 | 89.16 | 14.23 |
| CERN [20] | 86.84 | 84.83 | 88.17 | 1.88 |
| FG-AGR [46] | 88.38 | 87.52 | 91.09 | 2.31 |
| AMP_NET [43] | 88.52 | 87.57 | - | - |
| VTFF [29] | 88.29 | 87.20 | 88.81 | 0.45 |
| LANMSFF | 86.92 | 84.68 | 86.96 | 1.13 |

*D. Visualization*

Fig. 6 utilizes Grad-CAM [47] tool to visualize the activation maps of our model. It shows that the model highlights salient facial features, including the eyes, mouth, and nose, even in multi-view scenarios.

*E. Ablation Study*

Table XII presents an ablation study that investigates the impact of the MassAtt and PWFS modules on robustness to multi-view poses. To this end, we evaluate our proposed model with four different configurations: the whole model, the whole model without PWFS, the whole model without MassAtt, and the whole model without both PWFS and MassAtt on KDEF dataset. The results clearly indicate the significance of these modules for the model's robustness in multi-view settings. We employ Var metric for this purpose, where a lower Var value corresponds to a more robust model. Specifically, the absence of each module leads to an increase in Var by 0.4 for PWFS and by 1.94 for MassAtt. In addition, the simultaneous removal of both modules results in a 1-point rise in Var and a decrease in overall accuracy. This illustrates that both the proposed PWFS and MassAtt modules contribute to improving the performance of LANMSFF on multi-view poses by removing weak features and concentrating on key ones, respectively.

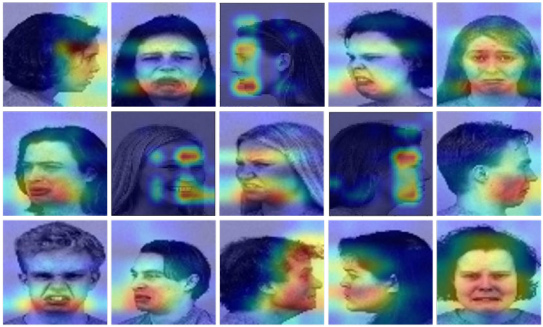

**Fig. 6.** Activation maps visualization using Grad-Cam. Salient regions are depicted in warm colors (e.g., red and yellow), while less important regions are represented with cold colors (e.g., blue and green).

TABLE XII
ABLATION STUDY ON TEST SET OF KDEF DATASET

| Methods | Acc (%) | | | | | | Var |
|---|---|---|---|---|---|---|---|
| | -90° | -45° | 0° | 45° | 90° | Overall | |
| Whole model | 89.44 | 91.18 | 92.04 | 91.00 | 90.17 | 90.77 | 0.66 |
| Whole model without PWFS | 89.85 | 91.28 | 92.05 | 92.84 | 90.16 | 91.24 | 1.05 |
| Whole model without MassAtt | 87.59 | 91.59 | 92.15 | 92.43 | 90.38 | 90.83 | 2.59 |
| Whole model without MassAtt and PWFS | 89.14 | 90.57 | 91.75 | 92.53 | 88.93 | 90.58 | 1.65 |

## V. CONCLUSION

This article presented LANMSFF to mitigate the challenges of high computational complexity and multi-view variations by suggesting a lightweight network including MassAtt and PWFS blocks to tackle multi-view head poses. We employ MassAtt block as an attention mechanism which generated spatial and channel attention maps to emphasize relevant features. Furthermore, PWFS module as a multi-scale feature selector was utilized to remove weak features before multi-scale feature fusion, unlike direct fusion in many existing methods. Experiments were conducted on KDEF, FER-2013, and FERPlus datasets. The results demonstrate that our model is robust to head-pose variation and achieves accuracy levels comparable to existing methods, using fewer parameters. Our future research will focus on dynamic in-the-wild datasets, incorporating spatiotemporal samples. In addition, we plan to incorporate pose estimation as a supplementary task.